\documentclass{article}

\usepackage{microtype}
\usepackage[table]{xcolor}
\usepackage{graphicx}
\usepackage{subcaption}
\usepackage{cuted}
\usepackage{capt-of}
\usepackage{booktabs} 
\usepackage{enumitem}
\usepackage{hyperref}

\usepackage{algorithm}
\usepackage{algpseudocode} 

\usepackage[preprint]{icml2026}

\usepackage{amsmath}
\usepackage{amssymb}
\usepackage{mathtools}
\usepackage{amsthm}
\usepackage{multirow}

\usepackage[capitalize,noabbrev]{cleveref}

\theoremstyle{plain}

\theoremstyle{definition}

\theoremstyle{remark}

\newcommand{\cut}[1]{}

\usepackage[textsize=tiny]{todonotes}

\icmltitlerunning{Out-of-Distribution (OOD) Detectors for Open-Set Radio-Frequency (RF) Fingerprinting}

\begin{document}

\twocolumn[
  \icmltitle{Out-of-Distribution (OOD) Detectors for Open-Set RF Fingerprinting}



  \icmlsetsymbol{equal}{*}

  \begin{icmlauthorlist}
    \icmlauthor{Sudeepta Mondal}{rtx}
    \icmlauthor{Ganesh Sundaramoorthi}{rtx}
  \end{icmlauthorlist}

  \icmlaffiliation{rtx}{RTX Technology Research Center (RTRC)}

\icmlcorrespondingauthor{Ganesh Sundaramoorthi}{ganesh.sundaramoorthi@rtx.com}

  \icmlkeywords{Machine Learning, ICML}

  \vskip 0.3in
]



\printAffiliationsAndNotice{}  


\begin{abstract}
Radio-frequency (RF) fingerprinting systems must operate in open-world environments where signals from unknown transmitters and temporal drift introduce distribution shift at test time. Out-of-distribution (OOD) detection provides a natural framework for this problem, yet its application to RF fingerprinting (RFF) remains limited. A key barrier to their adoption is that most OOD detectors require auxiliary OOD data for parameter tuning, an assumption that is difficult to satisfy in RF environments where representative OOD data is impractical to collect. In this work, we introduce a promising set of OOD detection methods from the machine learning literature to open-set RFF domain.  We present these methods within a unified mathematical framework based on information theory, which is a natural framework for communication systems. Our framework allows for the systematic analysis of methods and development of new methods. We further demonstrate the applicability of recent work on tuning OOD detectors without given OOD tuning data for open-set RFF. We evaluate on the POWDER RF fingerprinting dataset, showing that detectors tuned without any given OOD data achieve performance comparable to baselines with access to true OOD tuning data and greatly out-perform baseline approaches without access to true OOD tuning data, showcasing the practical viability for the RFF problem.
\vspace{-3mm}
\end{abstract}


\section{Introduction}

RF fingerprinting aims to identify wireless emitters from subtle signal characteristics introduced by hardware imperfections in the transmitter chain. These device-specific characteristics can help distinguish transmitters beyond protocol-level identifiers, making RF fingerprinting useful for wireless authentication, spectrum monitoring, and detecting unauthorized transmitters. Deep learning has significantly improved RF fingerprinting by learning discriminative representations from raw or transformed in-phase and quadrature (IQ) measurements. Prior work has explored a range of input representations, network architectures, and channel robustness strategies to improve identification performance under realistic wireless conditions \cite{Hanna_deep_learning, baldini2019, youssef2017, agadakos2019}.The fundamental limitation of this body of work is its closed-set assumption: any transmitter absent from the training set can be confidently misclassified as one of the known classes~\cite{hanna_openset}, directly undermining the security requirements of an RF fingerprinting system. 

The problem of detecting inputs from unknown classes is well-studied and active area in the machine learning community, where a rich body of out-of-distribution (OOD) detection methods has been developed, including confidence-based scores \cite{hendrycks2018, odin_liang}, energy-based metrics \cite{liu2021energy}, distance-based measures \cite{lee2018simple, sun2022knn}, and feature-shaping approaches \cite{sun2021reactoutofdistributiondetectionrectified, djurisic2022ash, xu2023vravariationalrectifiedactivation, mondal2025variationalinformationtheoreticapproach} that have shown promise across vision benchmarks \cite{yang2022generalized, zhang2024openoodv15enhancedbenchmark}, but the problem still remains a challenge as witenessed by benchmark results on OpenOOD \cite{zhang2023openood}, where methods have challenges generalizing across datasets. 

In contrast, open-set recognition in RF fingerprinting remains largely unexplored, with existing work limited to a small number of studies \cite{gritsenko2019, hanna_openset, Hanna_deep_learning, karunaratne2021opensetrffingerprinting}. These works establish that unknown transmitters can be detected by adapting open-set recognition architectures, yet reveal a consistent deployment challenge: performance is highly sensitive to the selection of transmitters in the authorized and outlier sets\cite{Hanna_deep_learning}. Leveraging the richer family of OOD detectors in this setting presents two key challenges. First, existing detectors are largely heuristic-driven \cite{sun2021reactoutofdistributiondetectionrectified, djurisic2022ash, xu2023vravariationalrectifiedactivation, zhao2024towards}, with limited guidance on method selection, though more principled information-theoretic formulations have recently been proposed \cite{mondal2025variationalinformationtheoreticapproach}. Second, many detectors require auxiliary OOD data to tune their parameters before deployment, an assumption that is particularly difficult to satisfy in RF environments where representative OOD data is impractical to collect in advance and detector performance can vary significantly with the choice of tuning data.

In this work, we introduce and examine the applicability of state-of-the-art OOD detection methods to Open-Set RF fingerprinting, including the practical challenge of detector tuning without any given OOD data. Our contributions are as follows.

\paragraph{Contributions:}
\vspace{-3mm}
\begin{enumerate}
    \item We introduce post-hoc OOD detection methods from the machine learning literature, with a focus on feature-shaping approaches, to the domain of RF fingerprinting within a unified mathematical framework whose information-theoretic underpinnings lend itself well to the communications community.
    
    \item We introduce recent work on OOD detector tuning without access to OOD data, addressing a key barrier to practical deployment, and demonstrate its applicability to open-set RF fingerprinting with state-of-the-art feature-shaping OOD detectors.

    \item We experimentally evaluate these methods on a publicly available RF fingerprinting dataset, demonstrating that effective open-set detection is achievable without representative OOD data, and establish a baseline for future work.

\end{enumerate}

\section{Related Work}

\subsection{Open Set RF Fingerprinting}

Deep learning has driven substantial progress in RF fingerprinting, with prior work establishing strong closed-set classification performance across a range of input representations and network architectures \cite{Hanna_deep_learning, baldini2019, youssef2017, agadakos2019}, yet this body of work does not generalize to unseen transmitters. A small number of works have explicitly addressed the open-set problem. Hanna et al. \cite{Hanna_deep_learning, hanna_openset} were among the first to adapt open-set recognition architectures to RF signals, revealing that performance is highly sensitive to transmitter set composition and temporal drift. \citet{gritsenko2019} demonstrated feasibility of new device detection without retraining, while \citet{karunaratne2021opensetrffingerprinting} proposed generative outlier augmentation to replace the need for real unauthorized transmitter data. More recent works have explored prototype and metric learning \cite{wang2023prototype, cai2025jrffp, ma2025mtpl}, and \citet{zhao2023ganrxa} address receiver impairment variability through a receiver-agnostic feature extraction framework. Overall, these works establish the feasibility of open-set RF fingerprinting but rely primarily on adapted open-set recognition architectures, leaving the richer family of OOD detection methods developed in the general ML community largely unexplored in this setting.

\subsection{OOD Detection Methods}

Early OOD detection methods relied on reconstruction-based autoencoders~\cite{autoencoders} and uncertainty-based approaches such as Monte Carlo dropout~\cite{gal2016dropout} and deep ensembles~\cite{lakshminarayanan2017simple}, but these are computationally expensive and often require training modifications~\cite{yang2022generalized}. Post-hoc methods, applied to pre-trained models without retraining, have since become a dominant paradigm \cite{yang2022generalized}, constructing scoring functions via confidence scores \cite{hendrycks2018, odin_liang}, energy-based metrics \cite{liu2021energy}, and distance-based measures \cite{lee2018simple, sun2022knn}.

A prominent family of post-hoc methods performs OOD detection by shaping intermediate network activations prior to scoring \cite{sun2021reactoutofdistributiondetectionrectified, Kong_BFAct, djurisic2022ash, fort2021exploring, zhao2024towards}. ReAct \cite{sun2021reactoutofdistributiondetectionrectified} clips large activation values, exploiting the observation that OOD inputs produce abnormally large activation spikes. ASH \cite{djurisic2022ash} sparsifies activations by flooring small values to zero with optional rescaling. More principled formulations in VRA \cite{xu2023vravariationalrectifiedactivation} and related works \cite{zhanglearning, mondal2025variationalinformationtheoreticapproach} derive shaping functions through optimization, moving beyond empirical heuristics. While predominantly studied in computer vision, these methods have also been applied in natural language processing \cite{jelenic_blood}, suggesting broader applicability across domains.

\section{Feature-Shaping OOD Detectors}

In this section, we introduce a subset of post-hoc OOD detectors in the machine learning literature, which have been referred to as \emph{feature shaping approaches}, within the context of the Open Set RF Fingerprinting (RFF) problem. We introduce these methods within a general mathematical framework to understand how one constructs them, the assumptions behind existing methods, and guide the development of new methods.

In the Open Set RFF problem, we are given a training set of signals from known distinct transmitters. In operation, the goal is to classify which transmitter emitted the received signal from the known transmitters or classify the signal as coming from an unknown transmitter.

\vspace{-2mm}
\subsection{What is OOD in RFF?} 

\begin{figure*}
    \centering
    \includegraphics[width=0.8\linewidth]{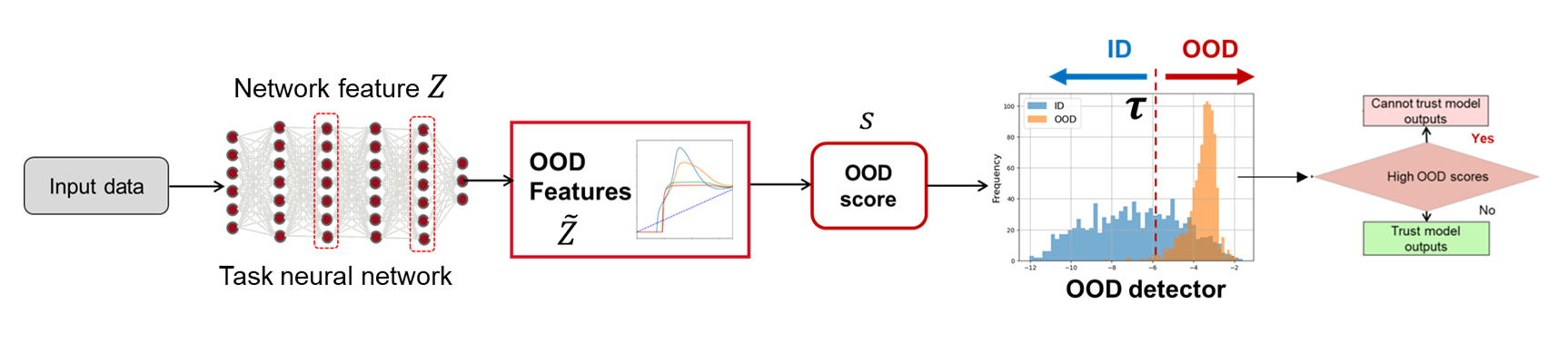}
    \caption{Feature Shaping OOD Detectors consist of two components that must be designed: the OOD feature $\tilde z = g_{\phi}(z)$ that processes task network feature $z$, and the OOD scoring function, $s=s(\tilde z)$. We propose the use of such OOD methods to the Open Set RFF problem.}
    \label{fig:OOD_Detection_Pipeline}
    \vspace{-6mm}
\end{figure*}
We frame Open Set RFF as an out-of-distribution (OOD) detection problem. In the OOD detection problem, a training set of known samples of different classes is given. In the context of neural networks, this data is used to train a task network (e.g., for the problem of classification). In operation, this network is used to classify incoming data. However, data in operation may be OOD or sampled from a different statistical distribution than the training set. OOD data may arise from new classes not seen in training, i.e., \emph{semantic shift}. In RFF, this can be data from a new transmitter. The network may give highly confident results on one of the known classes. Another source of OOD is \emph{covariate shifts}, i.e., variations in the data of known classes, e.g., rain or snow in images of known 'car' class, which may not be present in training data. Such covariate shift OOD may cause the network to give unreliable results, and should hence be marked as 'OOD' so the user of the system is alerted that the network cannot be trusted. In the context of RFF, covariate shifts could arise from signal variations from known transmitters such as new frequency modes not seen in training data or different variations in the signals due to different variations in the receivers. These cases could potentially result in unreliable results from the network, and hence these should also be flagged as 'OOD'.

\subsection{Problem Formulation}

We now present the formulation of the OOD detector problem, specifically feature shaping methods, within a unified information-theoretic framework, and indicate its application to Open Set RFF.

Given the trained task neural network (NN), $f_{\theta}$ where $\theta$ denotes the NN parameters, feature shaping OOD approaches consist of two components - OOD feature extraction and OOD scoring (Figure~\ref{fig:OOD_Detection_Pipeline}). Given input data, OOD feature extraction operates on task features computed from $f_{\theta}$ (e.g., in many cases the penultimate layer feature \cite{sun2021reactoutofdistributiondetectionrectified} but also intermediate layers \cite{djurisic2022ash}). We denote the task features selected from $f_{\theta}$ by $z$, i.e., this could be the penultimate layer feature or other intermediate layers. We denote by $g_{\phi}$ the OOD feature, which is a function of the task features, i.e., $\tilde z = g_{\phi}(z)$. Note the dependence of the OOD feature on parameters $\phi$. The goal in designing $g_{\phi}$ is to extract OOD relevant information from $z$. Although not mentioned in the literature, we postulate features $z$ of the network are extracted because they provide robustness to \emph{irrelevant change} in the data $x$. Note the data $x$ may contain many changes that still remain within the statistical distribution of the data used to train the network and within the generalization ability of the network (e.g., change of the background signal that while strictly different than training data may be within the generalization capability of the neural network); such changes do not constitute OOD and should be disregarded. The network $f_{\theta}$ consists of several layers of processing aiming to recover \emph{task} relevant features and in the process eliminating irrelevant changes, hence we postulate choosing $z$ may remove some irrelevant changes and thus is more robust to irrelevant change than using the data $x$ directly. The OOD feature computed from $z$ now shapes or tailors the feature from the original task of the network (e.g., classification) to be more relevant to the OOD task. After computing the OOD feature $\tilde{z}$, a scoring function $s$ maps $\tilde z$ to a scalar, and the OOD detector $d_{\phi}$ is defined as follows:
\begin{equation}
d_{\phi}(z) = 
\begin{cases}
    \textrm{ID} & s(g_{\phi}(z)) < \tau \\
    \textrm{OOD} &  s(g_{\phi}(z)) \geq \tau
\end{cases}.
\end{equation}

In the OOD problem, it is assumed that no OOD training/validation samples are available to determine $g_{\phi}$ and $s$ through training. In post-hoc feature shaping approaches, the trained network $f_{\theta}$ is given and it is assumed that the OOD detector does not change or modify the task network. In some methods, it is assumed that the ID training set or a subset is available, which we assume in our discussion.

\subsection{Review of Feature Shaping Methods}

In this subsection, we review and highlight some feature shaping approaches, including examples of various OOD features and scoring functions that yield state-of-the-art results on the OOD detection problem as measured by standard benchmarks in the machine learning community \cite{zhang2024openoodv15enhancedbenchmark}.

\begin{figure*}
    \centering
    \includegraphics[width=0.75\linewidth]{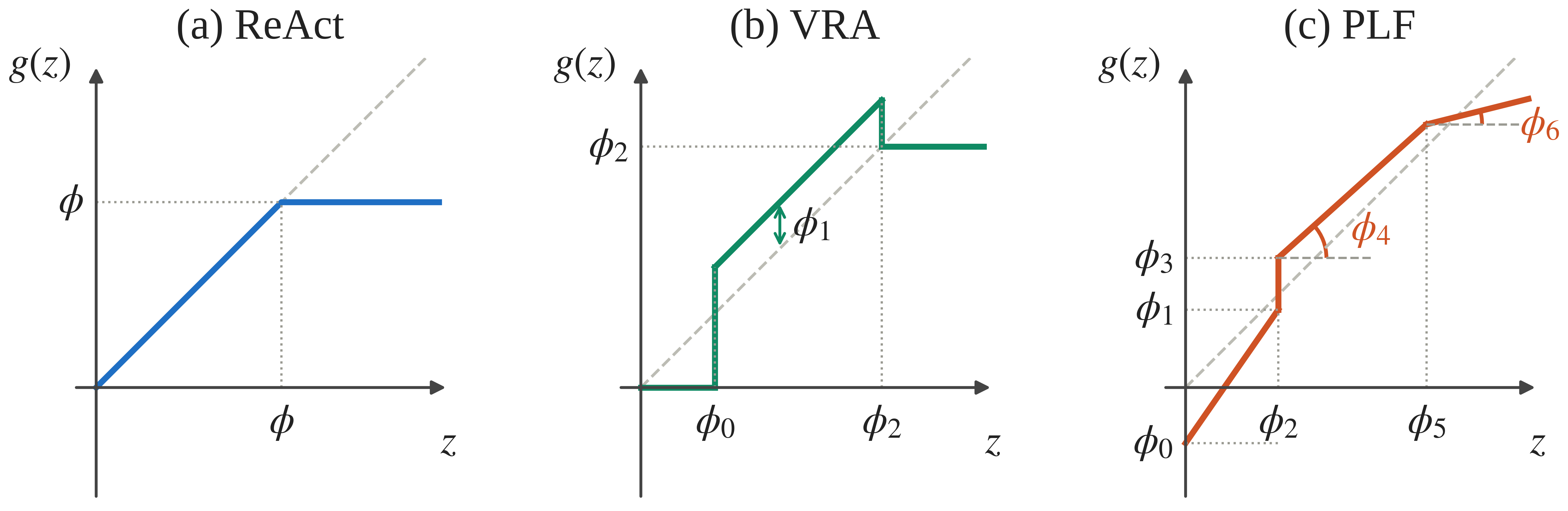}
    \caption{Representative shapes of the feature shaping methods discussed. (a) ReAct, (b) VRA and (c) PLF}
    \label{fig:shaping_functions_all}
\end{figure*}

First, we discuss two scoring approaches. We assume the classification task, which is closed set recognition in RFF. For this discussion, we assume $\tilde z = z$, i.e., the OOD feature is the identity map, and $z$ is the penultimate layer feature from the task NN. An obvious choice for the scoring function is the \emph{maximum softmax probability} (MSP), which is given as the maximum of the softmax probabilities 
\begin{equation}
    s_{\textrm{MSP}}(z) = \max_{c} p(c|z), \quad
    p(c|z) = \frac{\exp(W_cz+b_c)}{\sum_{c'} \exp(W_{c'}z+b_{c'})},
\end{equation}
where $W,b$ represents the weights associated with the last linear layer in the classification network. The idea is that if one of the class probabilities is high, then this suggests that the network has high confidence on a classification and this would indicate ID data. While intuitive, it has been shown that networks can achieve high probability (confidence) on OOD data and produce the wrong classification \cite{hendrycks2018,liu2021energy}. Accordingly, the \emph{energy score} was introduced \cite{liu2021energy}. The energy score is motivated by the Helmholtz free energy and is linked to the probability density of the data. The energy score is defined as follows:
\begin{equation}
    s_{\textrm{energy}}(z) = -T \log \sum_c \exp{\left[\frac 1 T (W_cz+b_c)\right]},
\end{equation}
where $T$ is a scaling parameter typically chosen as $T=1$. It is empirically shown that energy is higher for data with lower likelihood of occurrence, and it empirically leads to better separation between OOD and ID scores. Hence, many feature shaping approaches choose the energy score, and we will assume that in the rest of the discussion and experiments, although it is not universally used.

Next, we introduce three feature shaping approaches - ReAct \cite{sun2021reactoutofdistributiondetectionrectified}, variational rectification activation (VRA) \cite{xu2023vravariationalrectifiedactivation} and piecewise linear function (PLF) \cite{mondal2025variationalinformationtheoreticapproach} to highlight approaches in this area that achieve competitive performance. These approaches have considered $g_{\phi}$ to be a pointwise function of the network feature $z$, i.e., the same function is applied to each component of $z$ (typically $\geq 1000$ dimensions for deep networks) to form $\tilde{z}$. ReAct clips high values of $z$:
\begin{equation}
g_{\phi, \text{ReAct}}(z) = \min(z, \phi).
\end{equation}
Empirically, it was shown that OOD data results in spikes of the penultimate layer's features and that clipping these features produces a better separation between ID and OOD energy scores. No other motivation besides empirical observations are given.

Subsequently, VRA proposed a generalization of ReAct motivated by an optimization problem aiming to separate ID and OOD distributions by their mean value while keeping shaping close to the identity. By computations of the distributions of test OOD data on benchmark datasets, the following function was proposed:
\begin{equation}
    g_{\phi, \textrm{VRA}}(z) = 
    \begin{cases}
        0 & z < \phi_0 \\
        z +\phi_1 & \phi_0 \leq z < \phi_2 \\
        \phi_2 & z \geq \phi_2
    \end{cases},
\end{equation}
where the parameter set is $\phi=(\phi_0, \phi_1, \phi_2)$. In addition to clipping, it suppresses low values. This produced better results than ReAct on many datasets.

Another feature shaping approach is PLF \cite{mondal2025variationalinformationtheoreticapproach}, which proposes a seven-parameter family of piecewise linear shaping functions derived as an approximation to the solution of an optimization problem described in the next section. The PLF family is defined as follows:
\begin{equation}
    g_{\phi,\textrm{PLF}}(z) = 
    \begin{cases}
        \phi_0 + \frac{\phi_1-\phi_0}{\phi_2}z & 0\leq z < \phi_2 \\
        \phi_3 + \phi_4(z-\phi_2) & \phi_2 \leq z < \phi_5 \\
        \phi_3 + \phi_4(\phi_5-\phi_2) + & \\
        \phi_6(z-\phi_5) & z \geq \phi_5
    \end{cases},
\end{equation}
where $\phi = (\phi_0, \ldots, \phi_6)$ is the parameter set. PLF achieves state-of-the-art performance on standard benchmarks, outperforming VRA, ReAct, and most other feature-shaping approaches. Representative shapes of ReAct, VRA and PLF shaping functions are shown in Figure~\ref{fig:shaping_functions_all}. We can see the progression of shaping functions to more general ones, leading to better performance as witnessed through benchmark results in those papers. Note however, that more generality of the shaping function may not always lead to better OOD detection performance, due to overfitting on the tuning data, which could lead to poor generalization so care should exercised on the construction of such shaping functions.

\subsection{A Theoretical Framework for Feature Shaping}
While feature-shaping methods are often empirically motivated, notable attempts have been made to construct shaping functions in a systematic and principled manner through optimization \cite{zhao2024towards, xu2023vravariationalrectifiedactivation, mondal2025variationalinformationtheoreticapproach}. We summarize one such framework \cite{mondal2025variationalinformationtheoreticapproach}, which draws on information-theoretic tools originally developed for communications problems, making it of particular relevance to the wireless communication community. Furthermore, in addition to being a constructive framework, it also provides insights behind other existing methods, which is important to understand when such methods are expected to generalize.

Rather than considering a deterministic function $g_{\phi}$ to compute the OOD feature, a \emph{random feature} is considered, which provides a more general feature suitable for information theoretic methods. Using notation from probability theory, $Z$ is the task network feature treated as a random variable, and $\tilde Z$ denotes the random OOD feature. It is assumed that $\tilde Z \sim p(\tilde z|z)$ (similar to a channel in information theory \cite{cover1999elements}) and $Z\sim p(z|y)$ where $y\in \{\textrm{ID},\textrm{OOD}\}$; thus $Y \rightarrow Z \to \tilde Z$ forms a Markov Chain, where $Y$ is the random variable indicating OOD or ID. The data distribution $p(z|\textrm{ID})$ can usually be computed if the training set is available; however, $p(z|\textrm{OOD})$ is usually not known a-priori.

To show and understand many existing features arise from various distributional assumptions on ID and OOD data, \citet{mondal2025variationalinformationtheoreticapproach} studies various distributions of $Z$. To construct OOD features, equivalently, the distribution $p(\tilde z|z)$, and study them as a function of $p(z|y)$, an optimization formulation is proposed as follows:
\begin{equation}
\vspace{-0.3mm}
    L(p(\tilde z|z)) = -D_{KL}[p(\tilde z|\textrm{ID})||p(\tilde z|\textrm{OOD})] + \alpha 
    \mathrm{IB}[p(\tilde z|z)],
\end{equation}
where $D_{KL}$ denotes the symmetrized KL divergence, $IB$ denotes the \emph{information bottleneck} \cite{tishby2000information}, and $\alpha>0$ is a hyper-parameter. The goal is to minimize the loss over the set of probability distributions $p(\tilde z|z)$. The first term in the loss functional aims to maximize the distance as measured by KL divergence between the resulting distributions of $\tilde Z$ under the ID and OOD conditioning's so that the resulting OOD feature can separate ID and OOD data. The second term aims to construct the feature $\tilde Z$ to remove away data from $Z$ that is irrelevant to the task of OOD detection and keep only that which is relevant to OOD detection. Note that $Z$, being originally constructed for the task of the neural network, may contain data that is not relevant to the OOD task, and the goal is to remove such data, which could confuse the detector and cause false positives. The information bottleneck has been used widely in machine learning for various tasks, and is defined through mutual information as
\begin{equation}
    \textrm{IB}(p(\tilde z|z)) = I(\tilde Z;Z) -\beta I(\tilde Z;Y),
\end{equation}
where $\beta>0$ is a hyper-parameter. The first term minimizes the mutual information between $\tilde Z$ and $Z$, i.e., compresses $Z$, to remove data from $Z$ that may be from irrelevant change. The second term maximizes the mutual information between the OOD feature and $Y$, which is the random variable denoting whether the data is OOD or not. Thus, the latter term aims to retain information relevant to OOD.

An optimization algorithm was derived to compute the optimal OOD feature under various distributional assumptions under the simplification that the components of $Z$ and $\tilde Z$ are independent and each of the components of $Z$ and $\tilde Z$ have the same distribution, resulting in the optimization problem simplifying to the optimization of a one-dimensional distribution $p(\tilde z|z)$. To simplify matters further, that one-dimensional distribution was further considered to be Gaussian, i.e., $p(\tilde z|z)\sim \mathcal{N}(\mu(z),\sigma(z))$, for which an optimization scheme was formulated to determine the functions $\mu,\sigma$. Under the ID Gaussian assumption of the network feature and OOD Gaussian, Laplacian and Inverse Gaussian assumptions of the network feature, the mean value of the optimal features recover properties of existing features proposed in the literature. In fact, under Gaussian assumptions for ID and OOD, the optimal feature shape approximates ReAct. Under Gaussian ID and Laplacian OOD, the optimal shaping function resembles VRA. Thus, the theory appears to recover state of the art methods. PLF was proposed as an application of that theory to approximate the mean feature shapes arising from any one of the aforementioned distributional assumptions. Hence, PLF operates under more general distributional assumptions than ReAct and VRA, and other methods considered in the literature.

We note that the theoretical framework provides tools for not only understanding under what conditions OOD detectors can operate, but it provides a constructive framework for new methods. For example, one may want to model correlations between feature channels to generalize the assumptions under current approaches, and the above framework provides a methodological approach for incorporating such modeling assumptions. Given its roots in information theory and its ability to recover and predict state-of-the-art feature-shaping methods, this framework may prove useful to the wireless communications community for deriving new methods for open-set RFF with strong theoretical underpinnings.



\section{Approaches for Tuning OOD Detectors}

\begin{figure*}
    \centering
    \includegraphics[width=0.65\linewidth]{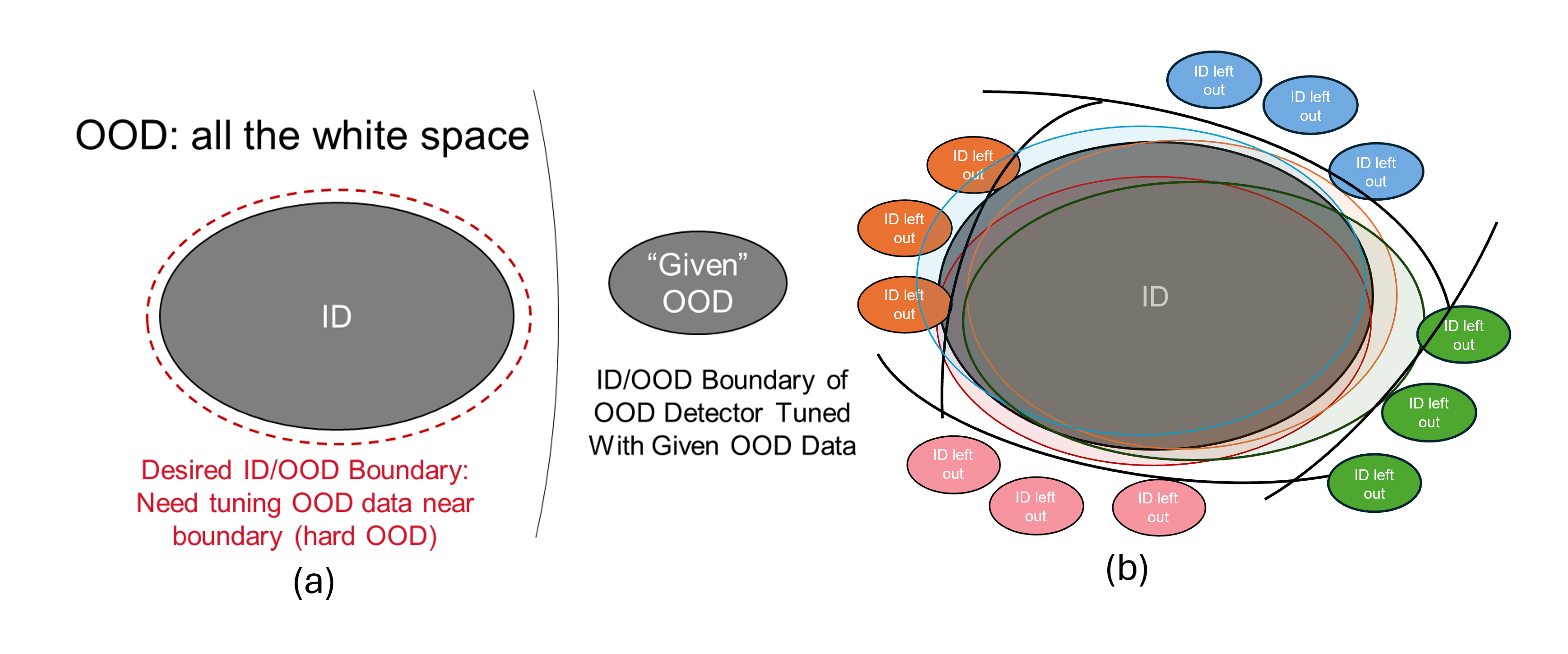}
    \caption{Intuition for SHOT approach proposed in~\cite{mondal2026tuningoutofdistributionooddetectors}: (a) A visual depiction of the space of all data.  The large ellipse indicates ID data.  The small ellipse indicates “given” OOD data, which may only cover a small part of all OOD. Tuning an OOD detector may result in ID/OOD boundary depicted by the black curve. If all true OOD data is to the right of the black curve, this tuned detector is adequate. An ideal boundary robust to any OOD would be the dashed red line. (b) A visual depiction of simulated hold-out dataset generation is shown, with 3 left out classes. Left-out classes are depicted in solid colors, and the same color transparent ellipse indicates the corresponding held-in classes. Held-in data may approximate true ID data (solid gray ellipse). The boundary (black solid curves) between one held out set and the held-in data may approximate a portion of the ideal ID/OOD boundary. By holding out multiple sets of random classes, SHOT approximates other parts of the ideal ID/OOD boundary.}
    \label{fig:intuition}
\end{figure*}


In order to deploy OOD detectors in practice, one would need to tune the parameters $\phi$ of the OOD detector $d_{\phi}$. This would require a tuning set of sample ID and \emph{OOD} data. In practice, it may be difficult obtain sample OOD data, which are often ``unknown unknown'' and this is also true in Open Set RFF. Benchmarks in ML such as OpenOOD \cite{zhang2024openoodv15enhancedbenchmark} may specify some held out data from the test OOD dataset for tuning, which not only may be unrealistic in practical settings, but OOD detectors are often sensitive to the choice of held out or tuning OOD data. This problem of tuning OOD detectors without given OOD tuning data has largely been neglected in the OOD literature until recently~\cite{mondal2026tuningoutofdistributionooddetectors}. We review some techniques for OOD detector tuning without a given OOD tuning set next. These approaches may provide a path to reliable deployment of OOD detectors in an operational setting, and hence may be ideally suited for the Open Set RFF problem.


We review and summarize three generic approaches to OOD detector tuning: Gaussian Noise Tuning, Adversarial Perturbations Tuning and Simulated Hold Out Tuning (SHOT).  The former two were considered baselines in and evaluated against SHOT in~\cite{mondal2026tuningoutofdistributionooddetectors}. SHOT performed generally the best overall on vision benchmarks, across both semantic and covariate shifts, and hence we adopt it in our experiments.

\textbf{Gaussian Noise:} Adopted by ReAct~\cite{sun2021reactoutofdistributiondetectionrectified} and VRA~\cite{xu2023vravariationalrectifiedactivation}, Gaussian noise images/signals are created by sampling $\mathcal{N}(\mu,\sigma)$ independently at each pixel/location~\cite{kirchheim2022pytorch}, serving as a natural OOD proxy since in-distribution data contains structure that noise does not.

\textbf{Adversarial Perturbations:} Adversarially perturbed ID data \cite{lee2018simple} serves as proxy OOD examples, as perturbations optimized via FGSM \cite{goodfellow2015explaining} shift samples into low-density regions of the data manifold, yielding lower likelihood under the training distribution. Adversarially perturbed ID data is formulated as follows: $x_{\mathrm{adv}} = x + \epsilon \cdot \mathrm{sign}\big(\nabla_x \mathcal{L}(x, c)\big)$, where $\epsilon$
is the norm of the perturbation, $\mathcal L$ is the loss for the task (e.g., cross-entropy for classification), and $c$ is the label for the data (e.g., class for the classification problem).

\textbf{Simulated Hold-Out Tuning:}
The crux of the method proposed in~\citet{mondal2026tuningoutofdistributionooddetectors} lies in observation that OOD data lies anywhere outside of the training (ID) distribution within some latent feature space (Figure~\ref{fig:intuition} (a)). Given a stationary training dataset, data samples held out from the training process may naturally be OOD to the task neural network trained on the held in data. As held out data is from the same training set, presumably collected using the same protocol and hence sampled from the same underlying distribution, the held out data may lie ``close'' to the held in distribution. Hence, this held-out data can be seen as ``near'' OOD examples. By tuning the parameters of a detector to discriminate between these ``near'' (OOD) examples and those of the ID distribution, SHOT hypothesizes that the OOD detector can identify both ``near'' and ``far'' OOD examples in novel datasets. As holding out a portion of data may only sample a portion of the ``near'' OOD space, SHOT considers multiple random choices of held out data to cover a larger portion of the ``near'' OOD space (Fig.~\ref{fig:intuition} (b)). To this end, SHOT simulates ID and OOD examples from the training dataset of the task NN, bypassing the need for a separate tuning dataset. SHOT creates a hold out dataset by placing all data belonging to randomly selected categories from the training dataset into a separate tuning dataset; note that any subset of the training data may be used so long as it is OOD to the held-in data -- held out categories clearly satisfy this property. 

Specifically, SHOT trains multiple task networks by holding out randomly chosen classes from each network. It then optimizes a loss function over the OOD detector parameters where the loss is measures the detectors loss in detecting held out data (i.e., simulated or proxy OOD data) from held in data over all trained network variants. The optimization is done through Bayesian optimization. Once the parameters are determined in training, these parameters are used for the OOD detector in conjunction with the original network (trained on all data). SHOT thus must train network variants offline, which adds non-negligible cost to training, however, this is often more convenient and reliable than collecting OOD tuning data, which depending on the application can involve significant engineering effort and multiple iterations of trial and error.

\vspace{-4mm}
\section{Experiments}

\subsection{Datasets and Settings}
\textbf{Dataset:} We evaluate on the POWDER RF fingerprinting dataset~\cite{reus_powder}, which contains over-the-air IQ recordings from four base stations deployed on the POWDER PAWR platform in Salt Lake City, Utah. The dataset was collected using USRP X310 base stations and a fixed USRP B210 receiver endpoint, capturing real-world channel effects across different transmitter locations and collection days. The transmitters emit standards-compliant waveforms with the data collection repeated over two independent days. In this work, we use the 4G LTE recordings from four base stations, \textit{bes}, \textit{browning}, \textit{honors}, and \textit{meb}, and construct open-set RF fingerprinting scenarios by treating one base station as OOD while training on the remaining base stations. The resulting dataset contains approximately 51{,}755 IQ slices per base station on the first collection day and 10{,}351 IQ slices per base station on the second collection day. 


\textbf{Pre-processing:} We scale the signals to each have unit energy by normalizing each signal by its $L_2$ norm.

\textbf{Models:} We use a one-dimensional ResNet-18 classifier as the in-distribution backbone for post-hoc OOD detection. The network involves one-dimensional convolutions over the temporal IQ sample dimension. Each input is a 512-sample complex IQ segment represented as a two-channel real-valued tensor containing the in-phase and quadrature components. The classifier is trained with cross-entropy loss to predict the transmitting base station from the in-distribution classes. For post-hoc OOD detection, we use the penultimate embedding extracted before the final linear classification layer.

\textbf{OOD Detectors:} We evaluate the SHOT framework of~\cite{mondal2026tuningoutofdistributionooddetectors} and their optimization settings on three post-hoc feature-shaping detectors: ReAct~\cite{sun2021reactoutofdistributiondetectionrectified}, VRA~\cite{xu2023vravariationalrectifiedactivation}, and PLF~\cite{mondal2025variationalinformationtheoreticapproach}, parameterized by one, three, and seven values respectively, applying increasingly expressive piecewise linear transformations to intermediate features prior to scoring. We use Energy score~\cite{liu2021energy} in conjunction with the feature shaping functions in our experiments, as proposed in the literature~\cite{sun2021reactoutofdistributiondetectionrectified,xu2023vravariationalrectifiedactivation,mondal2025variationalinformationtheoreticapproach}.


As a reconstruction error-based \textit{unsupervised baseline} OOD detector, we also train a ResNet-style 1D autoencoder using only in-distribution data. The encoder reuses the convolutional stem and four residual stages of the 1D ResNet-18 backbone, producing a latent temporal feature representation. The decoder consists of a sequence of transposed 1D convolutional layers that progressively upsample the encoded representation back to the original two-channel IQ segment. The autoencoder is trained to reconstruct the input IQ segment using mean-squared error loss. At test time, the OOD score is computed as the mean reconstruction error over both channels and all temporal samples, with larger reconstruction error indicating greater likelihood of being OOD.

\textbf{OOD Evaluation Scenarios:} We evaluate under three scenarios reflecting realistic open-set RF fingerprinting conditions. We treat each base-station/day pair as a distinct fingerprinting class. In all scenarios, the ID classifier is trained on six ID classes from three base stations, namely \textit{browning-day1}, \textit{browning-day2}, \textit{honors-day1}, \textit{honors-day2}, \textit{meb-day1}, and \textit{meb-day2}. The remaining base station, \textit{bes}, is held out as the true OOD transmitter and contributes two OOD classes, \textit{bes-day1} and \textit{bes-day2}, corresponding to different collection days and channel conditions. Scenario A tunes the known-OOD baselines on \textit{bes-day1} and evaluates on \textit{bes-day2}; Scenario B reverses this split by tuning on \textit{bes-day2} and evaluating on \textit{bes-day1}. Scenario C uniformly samples 10\%, 20\%, or 30\% of the available OOD data from both \textit{bes-day1} and \textit{bes-day2} for baseline tuning and uses the remaining OOD samples for evaluation. In all scenarios, the known-OOD baselines are granted access to a portion of the true OOD distribution for tuning, reflecting an optimistic assumption that is often impractical in real-world deployments. In contrast, SHOT requires no OOD data for tuning and relies only on in-distribution training data.

\begin{table*}[!t]
\centering
\scriptsize
\setlength{\tabcolsep}{3pt}
\begin{tabular}{l|cc|cc}
\toprule
 & \multicolumn{2}{c}{\textbf{Scenario A}} 
 & \multicolumn{2}{c}{\textbf{Scenario B}} \\
\cmidrule(lr){2-3} \cmidrule(lr){4-5}
Method 
& AU $\uparrow$ & FP $\downarrow$
& AU $\uparrow$ & FP $\downarrow$ \\
\midrule
ReAct (Baseline) 
& 0.534 & 0.655 
& 0.876 & 0.482 \\
\rowcolor{gray!20}
ReAct (SHOT)     
& 0.497 & 0.686 
& 0.931 & 0.295 \\
\hline
VRA (Baseline) 
& 0.706 & 0.567 
& 0.866 & 0.442 \\
\rowcolor{gray!20}
VRA (SHOT)     
& 0.696 & 0.574 
& 0.950 & 0.132 \\
\hline
PLF (Baseline) 
& 0.757 & 0.540 
& 0.450 & 0.995 \\
\rowcolor{gray!20}
PLF (SHOT)     
& 0.758 & 0.579 
& 0.932 & 0.172 \\
\bottomrule
\end{tabular}\hspace{0.3in}
%
\begin{tabular}{l|cc|cc|cc|cc|cc|cc}
\toprule
\multicolumn{13}{c}{\textbf{Scenario C}} \\
\midrule
 & \multicolumn{2}{c|}{\textbf{SHOT}}
 & \multicolumn{10}{c}{\textbf{Baseline: Known-OOD Tuning (\%)}} \\
\cmidrule(lr){2-3} \cmidrule(lr){4-13}
& &
& \multicolumn{2}{c|}{0.1\%}
& \multicolumn{2}{c|}{1\%}
& \multicolumn{2}{c|}{10\%}
& \multicolumn{2}{c|}{20\%}
& \multicolumn{2}{c}{30\%} \\
\cmidrule(lr){4-5}
\cmidrule(lr){6-7}
\cmidrule(lr){8-9}
\cmidrule(lr){10-11}
\cmidrule(lr){12-13}
 Method & AU $\uparrow$ & FP $\downarrow$
 & AU $\uparrow$ & FP $\downarrow$
 & AU $\uparrow$ & FP $\downarrow$
 & AU $\uparrow$ & FP $\downarrow$
 & AU $\uparrow$ & FP $\downarrow$
 & AU $\uparrow$ & FP $\downarrow$ \\
\midrule

ReAct
& 0.858 & 0.532
& 0.870 & 0.517
& 0.874 & 0.517
& 0.874 & 0.515
& 0.875 & 0.514
& 0.874 & 0.515 \\
\hline

VRA
& 0.907 & 0.364
& 0.889 & 0.448
& 0.919 & 0.394
& 0.926 & 0.357
& 0.925 & 0.359
& 0.925 & 0.356 \\
\hline

PLF
& 0.902 & 0.326
& 0.896 & 0.330
& 0.918 & 0.259
& 0.929 & 0.328
& 0.933 & 0.282
& 0.935 & 0.250 \\
\bottomrule
\end{tabular}
\caption{OOD detection performance comparison of SHOT~\cite{mondal2026tuningoutofdistributionooddetectors} with different baseline scenarios. Scenarios A \& B: comparison with baseline tuned on a single-day OOD partition. Scenario C: comparison with baseline tuned with a fraction of known OOD data used for baseline tuning. SHOT uses no OOD data at any stage of tuning.}
\label{tab:ood_all}
\vspace{-1mm}
\end{table*}

\textbf{SHOT Tuning:} We set the number of held-out categories to $M=2$ (approximately 33\% held out), consistent with the findings of~\citet{mondal2026tuningoutofdistributionooddetectors} that holding out 30--40\% of ID classes works well in practice. Note SHOT can automatically determine the optimal $M$ by evaluating a validation loss over held out data, and that results in $M=2$ on this data. The choice of $M=2$ results in 6 simulated ID/OOD splits out of the total number of ID classes, each defining a distinct held-out partition to cover a range of simulated ID/OOD splits (see Appendix for visualization), and thus we train 6 variant networks.

{\bf Bayesian Optimization:} Detector parameters are optimized via GP-based Bayesian optimization~\cite{frazier2018tutorialbayesianoptimization} with the \textit{GP-Hedge} acquisition function~\cite{Brochu_portfolio}, maximizing average AUROC over all simulated tuning sets following~\citet{mondal2026tuningoutofdistributionooddetectors}. The loss is the negative area-under the curve metric, evaluated as a sum over all 6 held-out network variants.

\subsection{Results}

We report AUROC (AU, $\uparrow$ higher is better) and FPR95 (FP, $\downarrow$ lower is better) as evaluation metrics~\cite{zhang2024openoodv15enhancedbenchmark}. Table~\ref{tab:ood_all} presents results for all scenarios. In Scenario A, regardless of the method of tuning we see that increasingly complex detectors perform better, that is, PLF outperform VRA which out performs ReAct. SHOT (in an unsupervised fashion) achieves competitive performance across all detectors, with the baseline (tuned on given OOD data) showing a modest advantage. However, the OOD detection performance is generally inferior in Scenario A (low AUROC, high FPR95). This suggests that \textit{bes-day2} is a challenging OOD evaluation distribution, possibly because its learned feature representation overlaps more strongly with the ID base stations. Although the known-OOD baselines are tuned using \textit{bes-day1}, this tuning provides only limited improvement. Since SHOT also performs similar to the baseline tuning in this setting, the degradation is likely driven by the intrinsic separability of the evaluation distribution rather than by the tuning strategy alone. 

In Scenario B, however, SHOT consistently outperforms the baseline (tuned using given OOD data), with notable AUROC gains for PLF (0.932 vs. 0.450) and VRA (0.950 vs. 0.866). The severe degradation of the PLF baseline in Scenario B is likely due to overfitting its higher-parameter family to the tuning data class, highlighting a key risk of supervised tuning with limited and potentially unrepresentative OOD data. SHOT, having no access to OOD data, avoids this issue and exhibits more consistent behavior across both scenarios. We note that in this scenario, the higher parameter detectors (VRA, PLF) out-perform significantly the lower parameter detector ReAct, with VRA slightly out-performing PLF, but comparable performance.

Table~\ref{tab:ood_all} also reports an ablation under Scenario C, where the baseline is tuned using varying fractions (0.1\%, 1\%, 10\%, 20\%, 30\%) of OOD data sampled uniformly across \textit{bes-day1} and \textit{bes-day2}. Across all detectors, performance improves when increasing the amount of known OOD tuning data from 0.1\% to 1\%, but shows little additional benefit beyond 10\%, with the most notable impact observed in the FPR95 reduction for PLF, the highest-parameter detector. This suggests that higher-parameter shaping methods are more sensitive to the quantity of OOD calibration data, while simpler detectors such as ReAct remain largely unaffected. SHOT performs comparably to baselines tuned with up to 30\% of the test OOD distribution across all three detectors, despite having no access to OOD data at any stage, demonstrating that effective detector tuning is achievable purely from in-distribution data and that the performance gap introduced by the absence of OOD data is marginal. Furthermore, higher parameter families out-perform the lower parameter ReAct, and both VRA and PLF perform about the same on the AU metric, but PLF out-performs VRA on the FP metric.

Table~\ref{tab:ood_unsupervised} compares SHOT-tuned feature-shaping detectors against the autoencoder baseline under fully unsupervised conditions, where neither method has access to any OOD data. The autoencoder performs poorly (AUROC 0.467), while all three SHOT-tuned OOD detectors substantially outperform the autoencoder, with VRA and PLF achieving AUROC above 0.90, demonstrating that post-hoc feature-shaping methods can be promising candidates for unsupervised open-set RF fingerprinting.

{\centering
\small
\setlength{\tabcolsep}{3pt}
\begin{tabular}{lcc}
\toprule
\textbf{Method} & \textbf{AU $\uparrow$} & \textbf{FP $\downarrow$} \\
\midrule
Autoencoder  & 0.467 & 0.747 \\
ReAct (SHOT) & 0.858 & 0.535 \\
VRA (SHOT)   & 0.907 & 0.371 \\
PLF (SHOT)   & 0.902 & 0.332 \\
\bottomrule
\end{tabular}
\captionof{table}{Fully unsupervised OOD detection with no tuning data: SHOT-tuned feature-shaping detectors vs. autoencoder baseline}
\label{tab:ood_unsupervised}
\par}
\vspace{-1mm}

\vspace{-3.0mm}
\section{Conclusion}
We proposed the use of post-hoc OOD detectors for the Open Set RF Fingerprinting problem. In particular, we presented feature-shaping approaches within the OOD detection literature as a viable set of approaches for Open Set RFF.  We presented these techniques within a unified information-theoretic framework, which provides a systematic framework to design new OOD detectors. Its information-theoretic underpinnings further provides a natural framework for integrating domain specific information from communication theory/principles. Experiments demonstrated the applicability of state-of-the-art OOD detectors to Open Set RFF, generally showing that higher parameter detector families out-performing lower parameter detectors if tuned with SHOT. We further reviewed different approaches for OOD detector tuning without access to representative OOD data, and demonstrated the applicability of a state-of-the-art tuning strategy (SHOT) to the open-set RFF problem. Experiments on the POWDER RF dataset under three realistic open-set scenarios show that SHOT-tuned feature-shaping detectors achieve performance comparable to detectors with access to true OOD tuning data, and substantially outperform reconstruction-based autoencoder baselines under fully unsupervised conditions. These results highlight the practicality of post-hoc OOD detection for open-set RF fingerprinting, where collecting representative OOD data is often infeasible, and establish a baseline for future work in this area, including extension to a broader family of OOD detectors, larger and more diverse RF datasets, and more challenging open-world deployment conditions.

\section*{Acknowledgments}
We thank Andrew Radlbeck for training the baseline classifier used in closed set classification and help in reviewing the RFF literature.

\comment{
\section*{Impact Statement}
This paper presents work whose goal is to advance the field of Machine Learning in Wireless Communications. There are many potential societal consequences 
of our work, none which we feel must be specifically highlighted here.
}


\bibliography{ood_lit}
\bibliographystyle{icml2026}

\newpage
\appendix
\onecolumn
\section*{Appendix}

\section{Visualization of the features}
\begin{figure}[h]
\centering
\begin{subfigure}[b]{0.45\linewidth}
    \centering
    \includegraphics[width=\linewidth]{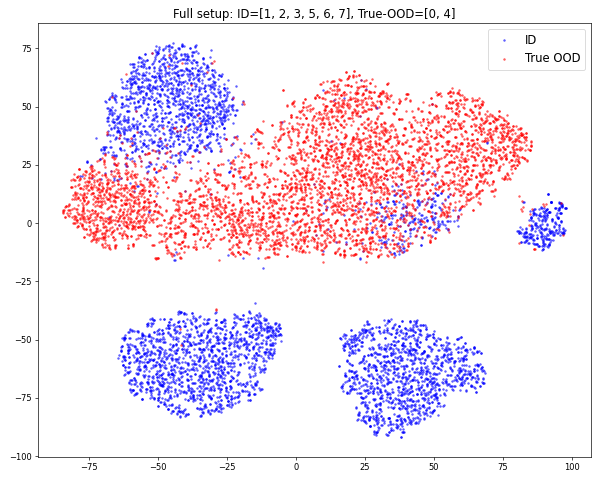}
    \caption{}
    \label{fig:sub1}
\end{subfigure}
\hfill
\begin{subfigure}[b]{0.45\linewidth}
    \centering
    \includegraphics[width=\linewidth]{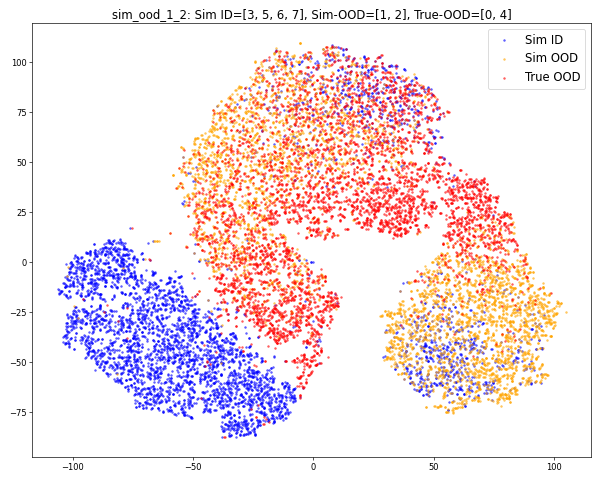}
    \caption{}
    \label{fig:sub2}
\end{subfigure}

\vspace{0.5em}

\begin{subfigure}[b]{0.45\linewidth}
    \centering
    \includegraphics[width=\linewidth]{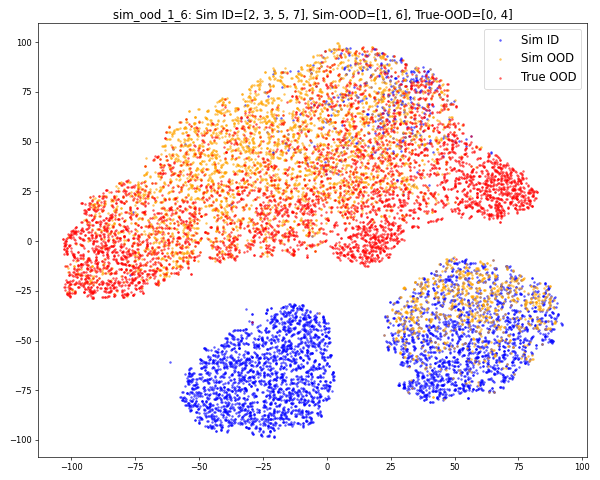}
    \caption{}
    \label{fig:sub3}
\end{subfigure}
\hfill
\begin{subfigure}[b]{0.45\linewidth}
    \centering
    \includegraphics[width=\linewidth]{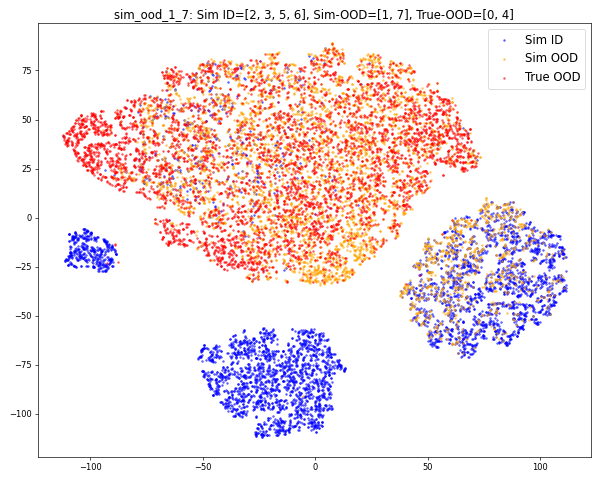}
    \caption{}
    \label{fig:sub4}
\end{subfigure}

\caption{t-SNE visualizations of ResNet-1D classifier penultimate layer features for the POWDER RF fingerprinting (a) shows the full ID classifier features in blue with true OOD classes \textit{bes-day1} and \textit{bes-day2} overlaid in red. (b)--(d) show three different random simulated ID/OOD splits using $M=2$ held-out classes, each with a different random seed. Blue denotes simulated ID (left-in) features, orange denotes simulated OOD (left-out) features, and red denotes true OOD features. The simulated OOD (orange) lies near the true ID/OOD boundary for majority of the simulated cases, and the union of orange points approximate a large portion of the boundary, providing empirical support in the RF fingerprinting example for the SHOT tuning intuition illustrated in Figure~\ref{fig:intuition}.}
\label{fig:main_figure}
\end{figure}

\end{document}